\documentclass[conference]{IEEEtran}
\IEEEoverridecommandlockouts
\usepackage{cite}
\usepackage{amsmath,amssymb,amsfonts}
\usepackage{algorithm}
\usepackage{algorithmic}
\usepackage{booktabs}
\usepackage{graphicx}
\usepackage{pifont}
\usepackage{textcomp}
\usepackage{xcolor}
\usepackage{hyperref}
\newcommand{\cmark}{\ding{51}}
\newcommand{\xmark}{\ding{55}}
\def\BibTeX{{\rm B\kern-.05em{\sc i\kern-.025em b}\kern-.08em
    T\kern-.1667em\lower.7ex\hbox{E}\kern-.125emX}}
\begin{document}

\title{Timestep-Aware SVDQuant-GPTQ for W4A4 Quantization of Wan2.2-I2V}

\author{
\IEEEauthorblockN{
Junhao Wu,
Dezhong Yao,
Hai Jin
}
\IEEEauthorblockA{
\textit{National Engineering Research Center for Big Data Technology and System} \\
\textit{Services Computing Technology and System Lab, Cluster and Grid Computing Lab} \\
\textit{School of Computer Science and Technology, Huazhong University of Science and Technology}\\
Wuhan, Hubei, China 430074\\
Email: \{junhao\_wu, dyao, hjin\}@hust.edu.cn}

}

\maketitle

\begin{abstract}
W4A4 quantization of large video diffusion Transformers offers substantial memory savings but is hindered by two main challenges: sparse large-magnitude activation outliers, and strongly timestep-dependent activation distributions across the multi-step denoising trajectory. These difficulties are compounded by Wan2.2-I2V's two-expert Mixture-of-Experts DiT design, whose high-noise and low-noise experts exhibit distinct quantization sensitivities that a single global calibration policy cannot capture. We propose a post-training quantization framework combining SVDQuant-based low-rank outlier compensation, GPTQ-based reconstruction-aware residual weight quantization, and timestep-bin-wise per-layer activation clipping-ratio search conducted independently for each expert. On the OpenS2V-Eval benchmark, our method reduces peak GPU memory by 59.3\% relative to the BF16 baseline while incurring only a 0.9\% drop in VBench average score and a 2.3\% drop in Imaging Quality, demonstrating that expert- and timestep-aware calibration is essential for high-fidelity W4A4 inference on MoE video DiTs.
\end{abstract}


\section{Introduction}
\label{sec:intro}

The rapid advancement of video generation has been driven largely by diffusion Transformer (DiT)\cite{peebles2023scalable} architectures operating at billion-parameter scale. Models such as Wan2.2 \cite{wan2025wan}, developed by the Alibaba team, represent the current frontier of open-source video generation: the Wan2.2-I2V-A14B model achieves state-of-the-art performance on public image-to-video benchmarks, yet its native inference requires a minimum of 80 GB of GPU memory on a single device. Community deployment pipelines such as ComfyUI routinely rely on W8A16 quantization to reduce this burden, but still demand high-end hardware. Pushing quantization further to W4A4 enables deployment on a wider range of devices.

The most challenging component of W4A4 quantization is 4-bit activation quantization. Unlike weights, activations are input-dependent and dynamically generated during inference. They often contain sparse but large-magnitude outliers in specific channels, which can dominate the quantization range. As a result, the majority of normal activation values are represented with very coarse resolution, causing the loss of fine-grained visual information. This effect is particularly harmful for video generation models, where activation errors can accumulate through the denoising trajectory and manifest as degraded textures, unstable boundaries, temporal flickering, or reduced subject consistency.

\begin{figure}[t]
    \centering
    \includegraphics[width=0.95\linewidth]{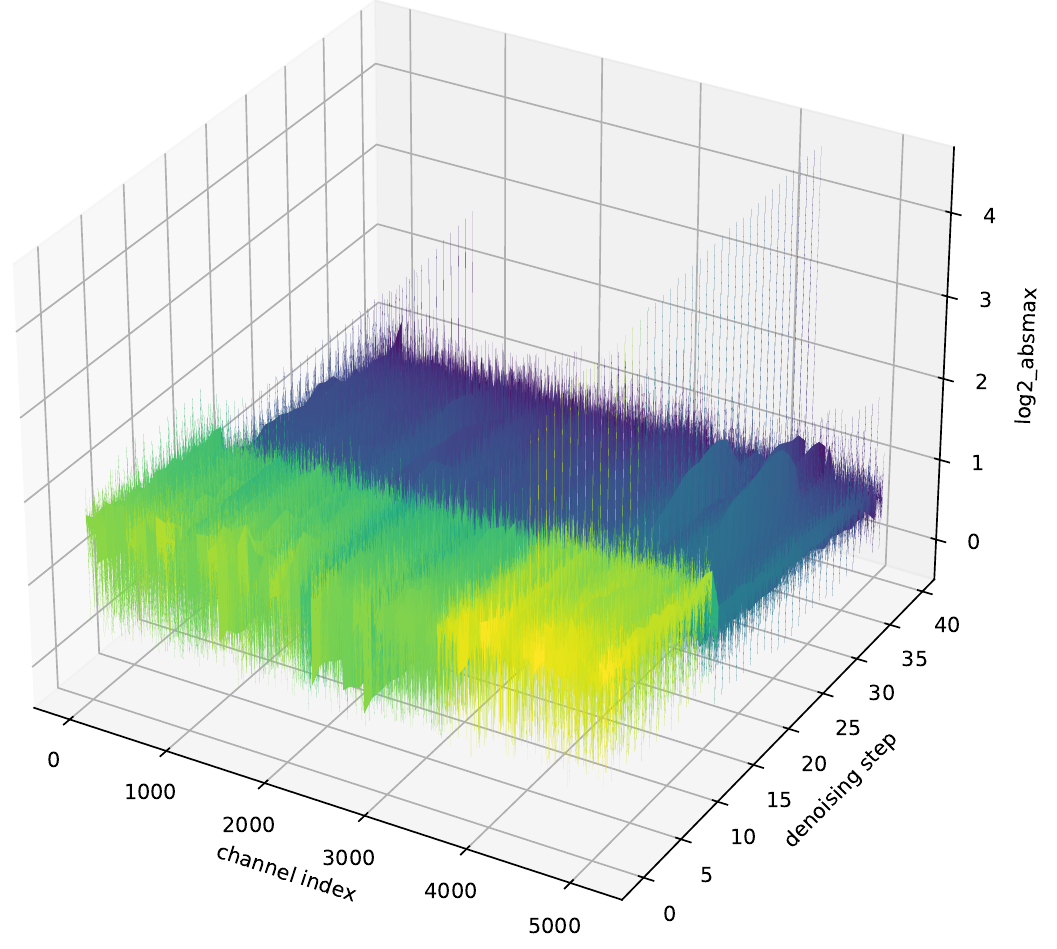}
    \caption{
    Activation non-stationarity across denoising timesteps in Wan I2V.
    We record the BF16 activation of \texttt{blocks.20.ffn.2} during a full image-to-video denoising trajectory and compute the per-channel statistic at each denoising step.
    }
    \label{fig:wan_i2v_activation_nonstationarity}
\end{figure}

Wan2.2 further amplifies this activation quantization difficulty due to two structural properties.  First, Wan2.2 adopts a two-expert Mixture-of-Experts DiT design, where a high-noise expert governs early denoising steps (global layout, coarse structure, large-scale spatiotemporal composition) and a low-noise expert governs later steps (texture refinement, boundary sharpness, visual convergence). These two experts have different activation statistics. As our block-wise analysis shows, the low-noise expert is systematically more fragile, especially in its attention output projections and FFN down projections. Applying a uniform quantization policy across both experts ignores this asymmetry and leads to avoidable quality loss. Second, activations in Wan2.2 are non-stationary across denoising timesteps, as shown in Fig.~\ref{fig:wan_i2v_activation_nonstationarity}. Early high-noise steps produce latent features with large dynamic ranges; later low-noise steps involve smaller, more structured activations where quantization error maps more directly onto perceptual degradation. Using a single set of activation quantization parameters for all timesteps forces a trade-off between over-clipping outliers in early steps and under-resolving fine details in late steps.

In this paper, we propose a W4A4 post-training quantization framework\footnote{The source code is available at \protect\url{https://github.com/CGCL-codes/Wan2.2-I2V-A14B-W4A4} and the quantized weights is available at \protect\url{https://huggingface.co/JunhaoWu/Wan2.2-I2V-A14B-W4A4}.} for Wan2.2-I2V that addresses these challenges. Our method integrates three components in a coherent calibration pipeline. First, we apply SVDQuant\cite{li2025svdquant} to decompose each quantized linear layer into a low-rank high-precision branch, and a residual branch that is quantized to 4 bits. Second, rather than rounding the residual branch weights to nearest, we apply GPTQ \cite{frantar2023gptq} to minimize layer-wise reconstruction error under the 4-bit constraint. GPTQ uses approximate second-order weight sensitivity information to reduce the residual quantization error that SVDQuant's low-rank branch does not absorb, providing a stronger main-branch quantization. Third, we perform timestep-bin-wise per-layer activation clipping-ratio search: the denoising trajectory is divided into coarse bins, and for each (layer, bin) pair, we select the clipping ratio that minimizes the layer-wise output reconstruction error on calibration samples. This allows the activation quantization range to adapt to the denoising stage with negligible runtime cost—only a lookup table indexed by the current timestep is needed at inference time. The same search is conducted independently for the high-noise and low-noise experts, naturally capturing expert-level differences.

We evaluate our method on the OpenS2V-Eval using the VBench evaluation protocol\cite{huang2024vbench}. Compared with the BF16 baseline, our W4A4 model reduces peak memory from 64.4 GB to 26.2 GB (a 59.3\% reduction) while achieving a VBench average score of 0.799, corresponding to only a 0.9\% degradation. Our main contributions are summarized as follows:

\begin{itemize}
    \item We identify and characterize two sources of W4A4 quantization difficulty specific to MoE video DiTs and show that both must be addressed jointly for high-fidelity quantization.
    \item We propose a hybrid W4A4 post-training quantization framework for video generation Transformers, combining SVDQuant-based low-rank compensation with GPTQ-based main-branch weight quantization.
    \item We introduce timestep-wise activation clipping-ratio search to handle the non-stationary activation distributions across denoising timesteps.
    \item On OpenS2V-Eval, our framework achieves a 59.3\% peak memory reduction with only a 0.9\% VBench average score degradation relative to BF16 inference.
\end{itemize}

\section{Related Work}

\subsection{Low-bit quantization for DiT}

General Transformer quantization has been extensively studied in large language models and vision Transformers. GPTQ~\cite{frantar2023gptq} formulates weight quantization as a layer-wise reconstruction problem and uses approximate second-order information to quantize large Transformer weights to 3 or 4 bits with small accuracy degradation. SmoothQuant~\cite{xiao2023smoothquant} observes that activation outliers are often more difficult to quantize than weights, and transfers quantization difficulty from activations to weights through an equivalent channel-wise scaling transformation. AWQ~\cite{lin2024awq} further shows that activation statistics can be used to identify salient weight channels and improve low-bit weight-only quantization. 

Recent studies have shown that DiT quantization introduces additional challenges beyond those in standard Transformers. PTQ4DiT~\cite{wu2024ptq4dit} identifies two major sources of quantization difficulty in DiTs: salient channels with extreme magnitudes and temporal variation of salient activations across denoising timesteps. To address these issues, it proposes Channel-wise Salience Balancing and Spearman's $\rho$-guided Salience Calibration, enabling effective W8A8 and W4A8 quantization for DiT models. Q-DiT~\cite{chen2025qdit} and TCAQ\cite{huang2025tcaq} further observes that DiTs exhibit significant spatial variance across input channels and temporal variance across timesteps. They introduce automatic quantization granularity allocation and sample-wise dynamic activation quantization, demonstrating that quantization parameters calibrated at one timestep may not generalize well to other timesteps. ViDiT-Q~\cite{zhao2025viditq} extends the analysis to both image and video DiT models, and designs a quantization scheme tailored for visual generation Transformers, achieving efficient W8A8 and W4A8 inference with dedicated GPU kernels. SVDQuant~\cite{li2025svdquant} targets 4-bit weight and activation quantization for diffusion models. Instead of only redistributing outliers between weights and activations as in smoothing-based methods, SVDQuant absorbs difficult-to-quantize outlier components with a high-precision low-rank branch. However, SVDQuant does not explicitly account for the timestep-dependent activation distribution shift during diffusion denoising.

\section{METHODOLOGY}

\subsection{W4A4 Linear Quantization}
We first formulate the W4A4 quantization problem for linear layers in Diffusion Transformers. In a Transformer block, most of the computation is dominated by linear projections in self-attention and feed-forward networks, including the query, key, value, output, gate and so on. Given an input activation matrix $X \in \mathbb{R}^{N \times C_{\mathrm{in}}}$ and a weight matrix $W \in \mathbb{R}^{C_{\mathrm{in}} \times C_{\mathrm{out}}}$, the full-precision linear operation is written as

\begin{equation}
    Y = XW,
\end{equation}

where $N$ denotes the number of tokens and $C_{\mathrm{in}}$, $C_{\mathrm{out}}$ are the input and output channel dimensions. In W4A4 quantization, both the weight matrix and the input activation are represented with 4-bit numerical formats during inference. The quantized linear operation can be written as

\begin{equation}
    \hat{Y} = Q_A(X) Q_W(W),
\end{equation}

where $Q_A(\cdot)$ and $Q_W(\cdot)$ denote the activation and weight quantizers, respectively. The goal of post-training quantization is to construct $Q_A$ and $Q_W$ without end-to-end retraining, such that the quantized output $\hat{Y}$ remains close to the full-precision output $Y$ under calibration data.

For a general uniform quantizer, the quantization and dequantization process can be expressed as

\begin{equation}
    Q(z; s, z_0) = s \cdot \left( \mathrm{clip}
    \left(
    \left\lfloor \frac{z}{s} \right\rceil + z_0,
    q_{\min}, q_{\max}
    \right) - z_0 \right),
\end{equation}

where $z$ is the tensor to be quantized, $s$ is the quantization scale, $z_0$ is the zero-point, and $[q_{\min}, q_{\max}]$ is the representable integer range. For symmetric quantization, the zero-point is set to zero, and the scale is computed as

\begin{equation}
    s = \frac{\alpha}{q_{\max}},
\end{equation}

where $\alpha$ denotes the clipping threshold. In 4-bit quantization, the representable range is very limited, making the choice of $\alpha$ particularly important. If $\alpha$ is too small, large-magnitude values are over-clipped; if $\alpha$ is too large, most normal values receive insufficient quantization resolution. This trade-off is especially critical for activations in diffusion models, whose distributions vary significantly across denoising timesteps.

\begin{figure*}[t]
    \centering
    \includegraphics[width=0.8\textwidth]{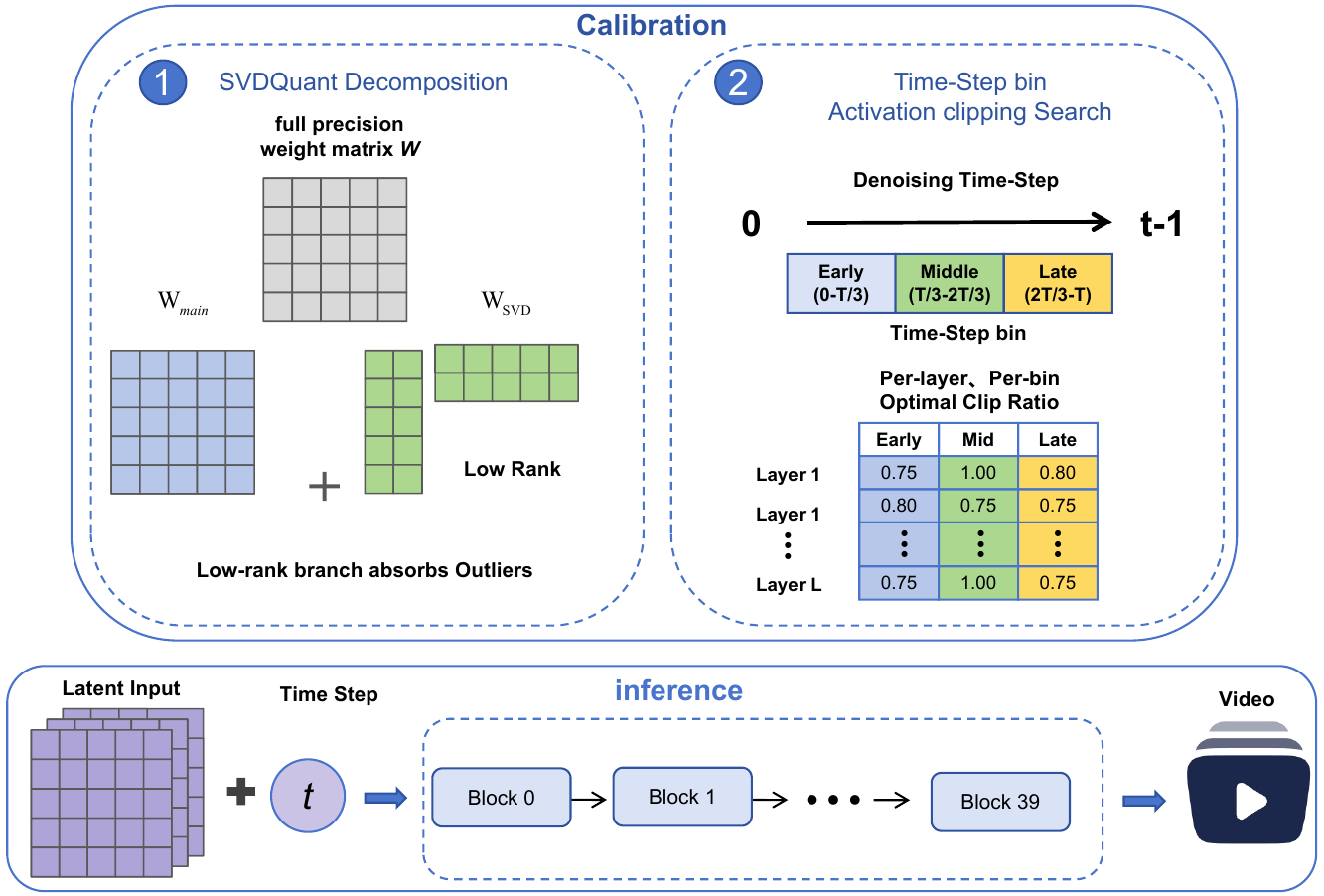}
\caption{
    Overview of the proposed W4A4 quantization framework.
During calibration, each sensitive linear layer is decomposed into a 4-bit main branch and a low-rank compensation branch by SVDQuant. Meanwhile, the denoising trajectory is divided into multiple timestep bins, and an activation clipping ratio is searched for each layer and timestep bin. During inference, the input latent and current timestep are processed by the quantized Transformer blocks using the precomputed clipping policy, producing the final generated video.
    }
    \label{fig:Framework}
\end{figure*}

\subsection{SVDQuant with GPTQ}

Among all components of W4A4 quantization, 4-bit activation quantization is usually the most fragile. Prior studies have shown that reconstruction-aware weight-only quantization can often preserve Transformer quality at 3--4 bits, while activation quantization is more difficult due to input-dependent distributions and sparse large-magnitude outliers~\cite{egiazarian2025bridging}. When activations are directly quantized to 4 bits, a few outliers can dominate the quantization range, causing most normal values to be represented with very coarse resolution. This is particularly harmful for video generation, where fine-grained activations are closely related to texture quality, object boundaries, and temporal consistency.

To reduce activation outliers, we adopt the key idea of SVDQuant. For a linear layer with activation $X_l$ and weight $W_l$, SVDQuant first applies a channel-wise smoothing transformation:
\begin{equation}
    X_l W_l = (X_l D_l^{-1})(D_l W_l)
    = \hat{X}_l \hat{W}_l ,
\end{equation}
where $D_l$ is a diagonal smoothing matrix. This transformation scales down outlier channels in the activation and transfers the corresponding magnitude to the weight. Since activations are dynamic while weights are static, this allows the difficult activation outlier problem to be handled offline on the weight side.

However, smoothing also magnifies outliers in the transformed weight $\hat{W}_l$, making direct 4-bit weight quantization inaccurate. Following SVDQuant, we decompose the transformed weight into a low-rank compensation branch and a residual branch:
\begin{equation}
    \hat{W}_l = L^l_1 L^l_2 + R_l .
\end{equation}
The low-rank branch $L^l_1 L^l_2$ is kept in higher precision to preserve outlier-sensitive directions, while the residual branch $R_l$ is quantized to 4 bits. In our implementation, we further apply GPTQ to the residual branch instead of using simple round-to-nearest quantization. The final quantized linear layer is computed as
\begin{equation}
    \hat{Y}_l
    =
    \hat{X}_l L^l_1 L^l_2
    +
    Q_A(\hat{X}_l)\hat{R}^{\mathrm{GPTQ}}_l .
\end{equation}
This design combines SVDQuant's outlier absorption with GPTQ's reconstruction-aware weight quantization. It provides a stronger W4A4 baseline than naive quantization, while leaving activation clipping as the remaining key factor to optimize.

\subsection{Timestep Activation Clipping Ratio Search}
After applying SVDQuant and GPTQ, the remaining major source of W4A4 degradation comes from activation quantization. In Wan2.2, activations are highly dependent on the denoising timestep. Early high-noise steps usually have larger dynamic ranges, while later low-noise steps are more sensitive to texture and detail preservation. Therefore, using a single activation clipping ratio for all timesteps can lead to sub-optimal quantization: a large range reduces the effective resolution of normal values, whereas a small range over-clips important outliers.

To address this issue, we perform timestep-wise activation clipping-ratio search. Specifically, we divide the denoising trajectory into several timestep bins. For each quantized layer and each timestep bin, we search the best clipping ratio from a small candidate set. Given a candidate clipping ratio $\rho$, the activation clipping threshold is computed as
\begin{equation}
    \alpha_{l,k}(\rho) = \rho \cdot m_{l,k},
\end{equation}
where $m_{l,k}$ denotes the reference activation magnitude of layer $l$ in timestep bin $k$, estimated from calibration samples. The corresponding activation scale is
\begin{equation}
    s^A_{l,k}(\rho) = \frac{\alpha_{l,k}(\rho)}{q_{\max}} .
\end{equation}

We select the clipping ratio by minimizing the layer-wise reconstruction error under the final quantized layer. For each layer $l$ and timestep bin $k$, the search objective is
\begin{equation}
    \rho^{*}_{l,k}
    =
    \arg\min_{\rho \in \mathcal{R}}
    \mathbb{E}_{t \in B_k}
    \left[
    \left\|
    Y_l(t) - \hat{Y}_l(t;\rho)
    \right\|_2^2
    \right],
\end{equation}
where $\mathcal{R}$ is the candidate set of clipping ratios, $B_k$ is the $k$-th timestep bin, $Y_l(t)$ is the full-precision layer output, and $\hat{Y}_l(t;\rho)$ is the output of the SVDQuant-GPTQ quantized layer using clipping ratio $\rho$.

During inference, no online search is required. For each denoising timestep, we simply map the current timestep to its corresponding bin and fetch the precomputed activation scale for each layer. The final quantized layer is computed as
\begin{equation}
    \hat{Y}_l(t)
    =
    \hat{X}_l L^l_1 L^l_2
    +
    Q_A(\hat{X}_l; s^A_{l,k})
    \hat{R}^{\mathrm{GPTQ}}_l,
    \quad t \in B_k .
\end{equation}
This simple lookup-based design allows activation quantization to adapt to the denoising stage with negligible runtime overhead. It also complements SVDQuant and GPTQ: SVDQuant reduces outlier-induced activation quantization difficulty, GPTQ optimizes the residual weight branch, and timestep-aware clipping further reduces activation error under W4A4 inference.

\section{EXPERIMENTS}

\subsection{Experiment Settings}

All experiments are conducted on the OpenS2V-Eval benchmark\cite{yuan2026opens2v}. The benchmark consists of 180 image-to-video samples covering eight categories, including single human, single object, single face, multiple humans, multiple objects, multiple faces, human-object interaction, and face-object interaction. We follow the VBench evaluation protocol\cite{huang2024vbench} and report four representative metrics: Imaging Quality, Aesthetic Quality, Subject Consistency, and Motion Smoothness. For all methods, we use the same generation configuration with $H \times W \times T = 720 \times 1280 \times 61$, where $H$, $W$, and $T$ denote video height, width, and number of frames, respectively.

\subsection{Main Results}

Table~\ref{tab:main_results} compares the proposed method with the full-precision baseline and several quantization baselines. We report both generation quality metrics and efficiency metrics to evaluate the quality-efficiency trade-off under the W4A4 setting.

\begin{table*}[t]
\centering
\caption{Main results on openS2V-eval. ``W'' and ``A'' denote weight and activation precision. ``TS-Clip'' indicates whether timestep-bin-wise activation clipping-ratio search is used. All 4-bit quantization uses the MXFP4 data format. Ours$^{\dagger}$ is a diagnostic variant that keeps the most sensitive \texttt{self\_attn.o} and \texttt{ffn.2} modules in higher precision, illustrating the impact of module-level sensitivity.}
\label{tab:main_results}
\resizebox{\textwidth}{!}{
\begin{tabular}{lccccccc}
\toprule 
Method 
& W 
& A 
& TS-Clip 
& VBench Avg. 
& Imaging Quality 
& Subject Consistency 
& Motion Smoothness  \\
\midrule
BF16 Baseline 
& BF16 
& BF16 
& -- 
& 0.806
& 0.705
& 0.739
& 0.975  \\

RTN
& 4-bit 
& 4-bit 
& \xmark 
& 0.733
& 0.389
& 0.861
& 0.949  \\

SVDQuant + RTN 
& 4-bit 
& 4-bit 
& \xmark 
&  0.783
&  0.671
&  0.705
&  0.974  \\

SVDQuant + GPTQ 
& 4-bit 
& 4-bit 
& \xmark 
&  0.793
&  0.684
&  0.722
&  0.974  \\

Ours 
& 4-bit 
& 4-bit 
& \cmark 
&  \textbf{0.799}
&  \textbf{0.689}
&  \textbf{0.733}
&  \textbf{0.975}  \\
\midrule
Ours$^{\dagger}$
& 4-bit 
& 4-bit 
& -- 
& 0.803
& 0.696
& 0.738
& 0.975  \\
\bottomrule
\end{tabular}
}
\end{table*}

\subsection{Sensitivity Analysis}

\begin{figure}[t]
    \centering
    \includegraphics[width=0.48\linewidth]{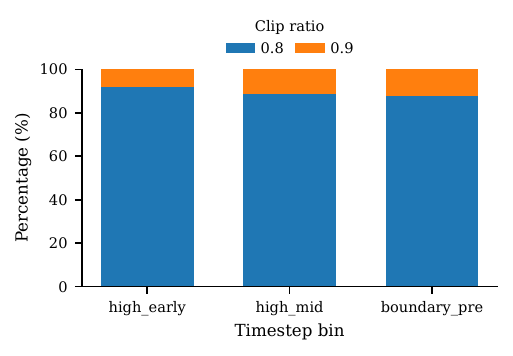}
    \includegraphics[width=0.48\linewidth]{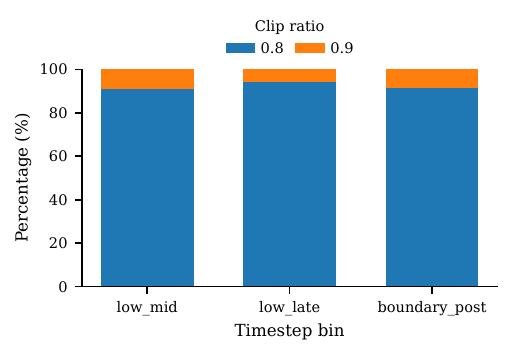}
\vspace{-0.5em}
    \caption{
    Distribution of selected timestep-bin-wise activation clipping ratios for the high-noise and low-noise experts.
    The two experts show different clipping preferences, supporting expert- and timestep-aware activation quantization.
    }
    \label{fig:clip_ratio_dist_both}
\vspace{-0.5em}
\end{figure}

\paragraph{Sensitivity analysis of timestep-aware clipping.}
Figure~\ref{fig:clip_ratio_dist_both} shows the distribution of selected activation clipping ratios across timestep bins for the high-noise and low-noise experts. For both experts, most modules prefer a clipping ratio of 0.8, while a smaller fraction selects 0.9. The ratio distribution varies across timestep bins and differs between the two experts, showing that a single global clipping ratio is sub-optimal for Wan2.2. These observations validate our timestep-aware clipping strategy, which adapts the activation quantization range to the denoising stage with negligible runtime overhead.

\begin{figure}[t]
    \centering
    \includegraphics[width=0.4\textwidth]{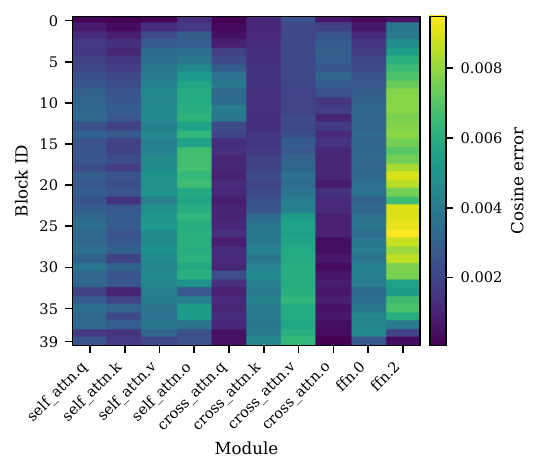}
    \hfill
    \includegraphics[width=0.4\textwidth]{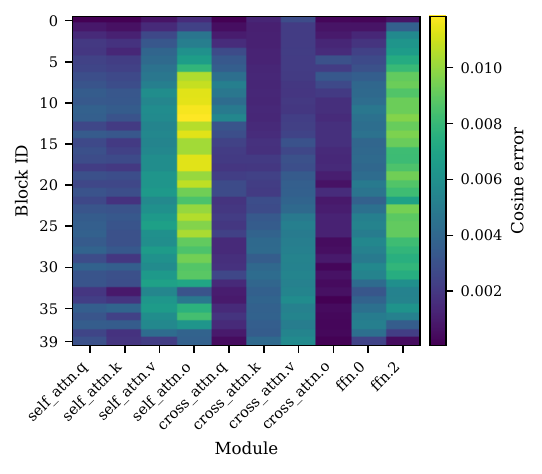}
    \caption{
    Cosine reconstruction error of different linear modules across Transformer blocks for the high-noise and low-noise experts.
    Brighter colors indicate larger quantization error.
    The most sensitive modules are \texttt{self\_attn.o} and \texttt{ffn.2}, while \texttt{self\_attn.q}, \texttt{self\_attn.k}, and \texttt{cross\_attn.o} are relatively robust.
    The low-noise expert generally exhibits larger errors than the high-noise expert, especially in middle and deeper blocks.
    }
    \label{fig:block_sensitivity}
\end{figure}

\paragraph{Sensitivity analysis of Transformer blocks.}
The challenge rule for the MXFP4 setting allows at most five Transformer blocks to be kept in higher precision. To better understand which parts of Wan2.2 are most sensitive to W4A4 quantization, we further analyze the block-wise and module-wise reconstruction error. Figure~\ref{fig:block_sensitivity} shows the cosine reconstruction error of different linear modules across Transformer blocks for the high-noise and low-noise experts.

Several clear patterns can be observed. First, the quantization sensitivity is highly module-dependent. Among all linear modules, \texttt{self\_attn.o} and \texttt{ffn.2} consistently exhibit the largest errors in both experts, indicating that the attention output projection and the FFN down projection are the most fragile components under W4A4 quantization. Second, the sensitivity also varies with model depth. In both experts, larger errors are mainly concentrated in the middle and deeper Transformer blocks, suggesting that these blocks are more vulnerable to quantization noise. Third, the low-noise expert is generally more sensitive than the high-noise expert, especially in \texttt{self\_attn.o} and \texttt{ffn.2}. This is consistent with the role of the low-noise expert in late-stage denoising, where texture refinement, local details, and visual convergence are more sensitive to numerical perturbations.

These observations provide guidance for high-precision retention. However, we find that keeping only a small number of complete Transformer blocks in higher precision brings limited improvement, since the dominant errors are not confined to a few isolated blocks but are distributed across specific module types in many blocks. As a diagnostic experiment, when we keep the \texttt{self\_attn.o} and \texttt{ffn.2} modules in both experts at higher precision, the generated video quality improves noticeably. This suggests that module-level sensitivity is more concentrated than block-level sensitivity, and that output projections and FFN down projections are the primary sources of W4A4 degradation in Wan2.2.

\subsection{Qualitative Results}

\begin{figure}[t]
    \centering
    \includegraphics[width=\linewidth]{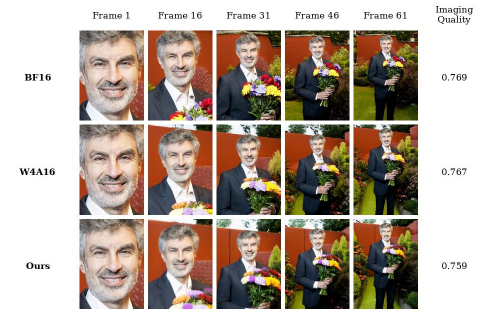}
    \caption{
    Qualitative comparison among the BF16 baseline, the W4A16 quantized model, and our W4A4 model.
We show frames 1, 16, 31, 46, and 61, together with the corresponding Imaging Quality score.
Our W4A4 model preserves the subject identity, pose, and scene structure with moderate visual degradation.
    }
    \label{fig:qualitative_comparison}
\end{figure}

Figure~\ref{fig:qualitative_comparison} presents a qualitative comparison among the BF16 baseline, the W4A16 quantized model, and our W4A4 model. We sample five frames from the generated video and report the corresponding Imaging Quality score for each method. The BF16 baseline achieves an Imaging Quality score of 0.769, while W4A16 obtains a very close score of 0.767. Our W4A4 model achieves 0.759, showing only a moderate drop despite the significantly more aggressive quantization setting.

Visually, our method preserves the main subject identity, pose, clothing, bouquet, and overall scene layout across the sampled frames. The generated frames remain temporally coherent from the close-up initial frame to the later full-body frames. Compared with BF16 and W4A16, our W4A4 result exhibits slightly stronger texture smoothing and minor detail degradation, especially around facial details and flower boundaries. However, the overall structure and semantic content are well maintained, and no severe artifacts or collapse are observed.

\subsection{Efficiency Results}

Table~\ref{tab:efficiency_results} reports the efficiency comparison among BF16, W4A16, and our W4A4 model. All results are measured on a single NVIDIA H100 GPU under the same generation setting. Compared with the BF16 baseline, W4A16 reduces the model size from 64.0 GB to 26.4 GB and the peak memory from 64.4 GB to 26.8 GB. Our method further reduces the model size to 25.9 GB and the peak memory to 26.2 GB, achieving a similar memory footprint to W4A16 while quantizing both weights and activations.

\begin{table}[t]
\centering
\caption{
Efficiency comparison of different precision settings.
Model size, peak memory, latency, and throughput are measured under the same inference configuration.
}
\label{tab:efficiency_results}
\resizebox{\linewidth}{!}{
\begin{tabular}{lcccc}
\toprule
Method & Model Size (GB) & Peak Memory (GB) & Latency (s) & Throughput (FPS) \\
\midrule
BF16  & 64.0 & 64.4 & 520.7 & 0.117 \\
W4A16 & 26.4 & 26.8 & 534.5 & 0.114 \\
Ours  & 25.9 & 26.2 & 955.1 & 0.064 \\
\bottomrule
\end{tabular}
}
\end{table}

The main benefit of our method is memory reduction. Compared with BF16, our W4A4 model reduces the model size by 59.5\% and peak memory by 59.3\%. This makes the model much easier to deploy on memory-constrained devices. However, our current implementation does not yet achieve latency improvement. The reason is that the H100 GPU used in our experiments does not provide a native FP4/MXFP4 execution path for this model, and our implementation relies on pseudo-quantization. Specifically, low-bit weights and activations are dequantized online before matrix multiplication, and the current dequantization path is implemented without a fused CUDA kernel. As a result, the Python-level dequantization and unfused operators introduce substantial overhead, increasing latency from 520.7 seconds to 955.1 seconds.

Therefore, the latency reported in Table~\ref{tab:efficiency_results} should be interpreted as the cost of our reference pseudo-quantized implementation rather than the theoretical speed of a fully optimized W4A4 deployment. With hardware support for native FP4/MXFP4 tensor operations or fused dequantization-GEMM kernels, the proposed quantization method is expected to better translate its reduced model size and activation precision into practical inference speedup.

\bibliographystyle{IEEEbib}
\bibliography{icme2026references}

\end{document}